\documentclass[runningheads]{llncs}
\usepackage[T1]{fontenc}
\usepackage{graphicx,verbatim}
\usepackage{tikz}
\usetikzlibrary{arrows.meta,positioning,calc}
\usepackage{todonotes}
\usepackage{color}

\usepackage{graphicx} % Required for inserting images
\usepackage{setspace}
\usepackage{soul}
\usepackage{amssymb}
\usepackage{amsmath}
\usepackage{pifont}
\usepackage{tikz}
\usepackage{tikz, forest}
\usepackage{xparse}
\usepackage{pgfplots, pgfplotstable}
\usepgfplotslibrary{fillbetween,statistics}
\pgfplotsset{compat=1.18}
\usepackage{algorithm}
\usepackage{algpseudocode}
\usepackage{filecontents}
\usepackage{multirow}
\usepackage{mathtools}
\usepackage{booktabs}
\usepackage{xcolor}
\usepackage{rotating} 
\usepackage{url}
\usepackage{hyperref}
\usepackage{array}   % for \newcolumntype macro
\usepackage[separate-uncertainty]{siunitx}
\usepackage{etoolbox}
\usepackage{caption}
\usepackage{subcaption}
\usepackage{tablefootnote}
\usepackage{transparent}
\renewrobustcmd{\bfseries}{\fontseries{b}\selectfont}
\newcolumntype{C}{\num{S}} % math-mode version of "l" column type

\definecolor{cbBlack}{HTML}{000000}
\definecolor{cbOrange}{HTML}{E69F00}
\definecolor{cbBlue}{HTML}{56B4E9}
\definecolor{cbGreen}{HTML}{009E73}

\definecolor{cbTol1}{HTML}{DDCC77}
\definecolor{cbTol2}{HTML}{CC6677}
\definecolor{cbTol3}{HTML}{AA4499}
\definecolor{cbTol4}{HTML}{882255}

\usepgfplotslibrary{fillbetween}
\usetikzlibrary{arrows.meta}
\usetikzlibrary{positioning}
\usetikzlibrary{shapes.geometric}   % For trapezium 
\usetikzlibrary{spy}
\usetikzlibrary{3d}
\usetikzlibrary{plotmarks}
\usetikzlibrary{shapes,arrows,positioning,calc}

\newcommand{\bx}{\boldsymbol{x}}

\newcommand{\ap}{\mathrm{AP}}
\newcommand{\ar}{\mathrm{AR}}
\newcommand{\apat}[1]{\mathrm{AP}_{#1}}

\newcommand{\added}[1]{\,{\fontsize{8}{9.5}\selectfont\textcolor{blue}{+#1}}}
\newcommand{\lost}[1]{\,{\fontsize{8}{9.5}\selectfont\textcolor{red}{-#1}}}
\renewcommand{\b}[1]{\boldsymbol{#1}}
\begin{document}
%
% \title{Diffuse what you know: Opportunistic labels in training and inference}
\title{Exemplar Diffusion: Improving Medical Object Detection with Opportunistic Labels}
\titlerunning{Exemplar Diffusion}
 %% Removed for anonymized MICCAI submission
\author{Victor Wåhlstrand\inst{1}\orcidID{0000-0001-6569-120X} \and
Jennifer Alvén\inst{1}\orcidID{0000-0003-4195-9325} \and
Ida Häggström\inst{1,2}\orcidID{0000-0001-9178-6683}}
%index{Wåhlstrand, Victor}
%index{Alvén, Jennifer}
%index{Häggström, Ida}
%
\authorrunning{V. Wåhlstrand et al.}
\institute{
    Chalmers University of Technology, SE-41296 Göteborg, Sweden
    \email{victor.wahlstrand@chalmers.se}
}

% \author{Anonymized Authors}  %% Added for anonymized MICCAI submission
% \authorrunning{Anonymized Author et al.}
% \institute{Anonymized Affiliations \\
% \email{email@anonymized.com}}
\maketitle              % typeset the header of the contribution
\begin{abstract}
We present a framework to take advantage of existing labels at inference, called \textit{exemplars}, in order to improve the performance of object detection in medical images. The method, \textit{exemplar diffusion}, leverages existing diffusion methods for object detection to enable a training-free approach to adding information of known bounding boxes at test time. We demonstrate that for medical image datasets with clear spatial structure, the method yields an across-the-board increase in average precision and recall, and a robustness to exemplar quality, enabling non-expert annotation. Moreover, we demonstrate how our method may also be used to quantify predictive uncertainty in diffusion detection methods. Source code and data splits openly available online: \url{https://github.com/waahlstrand/ExemplarDiffusion}

  % Authors must provide keywords and are not allowed to remove this Keyword section.

\end{abstract}
%
%
%

%% removed for anonymized MICCAI submission.

% The following acknowledgement and disclaimer sections can be removed for the double-blind review process.  If and when your paper is accepted, reinsert the acknowledgement and the disclaimer clause in your final camera-ready version.
% IF you opted to include the acknowledgement and disclaimer sections, they will count towards the 8-page limit.

\section{Introduction}

Objects in medical images are typically not randomly distributed but spatially correlated, and knowledge of one object's location may inform us about the location of others. In the context of object detection, the existence of partial annotations or minimal information provided by an expert should provide valuable information about nearby, complementary objects. We consider the setting where such partial annotations (e.g., bounding boxes) are available and can be leveraged to improve object detection in the same image during inference. Unlike interactive segmentation \cite{chenConditionalDiffusionInteractive2021}, the provided annotation is known and should aid in the detection of other objects.

Opportunistic test-time annotations may be provided by an expert, or already exist in the dataset and would otherwise not be used during inference. Due to the cost of full medical image annotation, both public and proprietary medical image datasets frequently contain partial and missing labels \cite{yanLearningMultipleDatasets2021,zhouPriorAwareNeuralNetwork2019,waahlstrand2024explainable}. While fully annotated open-source datasets have played an important role in computer vision and medical image analysis, providing carefully curated clinical data for developing and benchmarking detection models, many datasets are partially annotated or were originally compiled from heterogeneous sources \cite{xuCADSComprehensiveAnatomical2025,labella2024BrainTumor2026}.

Orthogonal approaches typically handle partial labels through additional manual annotation or semi‑supervised methods aimed at learning from incomplete annotations, such as pseudo‑labelling \cite{labella2024BrainTumor2026} or label propagation \cite{ganDeepSemisupervisedLearning2022}. These strategies, however, generally require retraining and are not designed to incorporate partial annotations during inference. Recently, Carion et al. introduced SAM3 \cite{carion2025sam3segmentconcepts}, demonstrating that guiding segmentation with \textit{exemplars} can be a powerful mechanism for discovering visually similar structures at inference. However, SAM3 uses a learnt encoder for exemplars and other prompts, requiring a substantial data engine and training \cite{kirillovSegmentAnything2023,carion2025sam3segmentconcepts}.
  
In the context of e.g. segmentation and detection, exemplars already exist as maps or coordinates in the image domain, and circumventing the need to learn an additional representation could lead to a more interpretable and less data-hungry method for adding annotation information during inference.

% is designed for zero-shot segmentation, and therefore lacks ways to constrain detections to clinically relevant object classes, without substantial retraining. 

We present \textit{exemplar diffusion}, a framework based on detection diffusion \cite{chenDiffusionDetDiffusionModel2023}, that leverages existing, opportunistic annotations in medical imaging to improve the detections of an already trained model. Our main contributions are four-fold: \textit{i}) a training-free approach for adding exemplars to a trained model that boosts detection performance, \textit{ii}) an embedding-free method of representing bounding box exemplars in detection, while also demonstrating that the method is \textit{iii}) noise insensitive, enabling quicker non-expert annotations, and \textit{iv}) may be used estimate the detection uncertainty using the known exemplars.

\subsection{Related work}
Modern object detection methods typically rely on dense proposal frameworks such as Faster R-CNN\,\cite{renFasterRCNNRealTime2016,albuquerqueDeepLearningbasedObject2025} or learnable query‑based approaches like DETR\,\cite{Carion2020} and its variants. Sparse R-CNN\,\cite{sunSparseRCNNEndtoEnd2021a} provides a hybrid alternative by learning a compact set of proposals directly in bounding‑box space.

DiffusionDet\,\cite{chenDiffusionDetDiffusionModel2023} reframes Sparse R-CNN’s proposal mechanism within a diffusion process\,\cite{hoDenoisingDiffusionProbabilistic2020,songDenoisingDiffusionImplicit2022}. Ground‑truth boxes are perturbed with noise during training, and a network is optimized to iteratively denoise them, enabling inference by refining randomly initialized boxes into plausible detections. In medical imaging, Hamamci et al.\,\cite{hamamciDiffusionBasedHierarchicalMultiLabel2023} apply DiffusionDet to model hierarchical dependencies in panoramic dental X‑rays, and others have also achieved competitive results in tumour cell detection\,\cite{liangImprovingCirculatingTumor2024}, histopathology analysis\,\cite{yangTSBPImprovingObject2024}, and lesion detection\,\cite{zhao_diffuld_2023}.

Reliable uncertainty estimation is crucial in medical image analysis, where automated predictions often inform high‑stakes clinical decisions. In object detection, uncertainty is often modelled by explicit variance networks \cite{harakehEstimatingEvaluatingRegression2021}, requiring explicit design choices before training. Post-hoc methods are typically ensemble methods, but diffusion models may also be used for this purpose. \cite{wintelUsingEnsembleDiffusion}
% SAM3 builds on the foundation of DETR, and represents objects and prompts as embedded query representations, the former using a sophisticated prompt encoder, enabling joint representation of several prompt modalities. The downside of this approach is that this representation must be learnt. 

% In DiffusionDet, bounding boxes are not embedded, but directly represented as coordinate vectors. Without any invasive changes, we adapt the DiffusionDet method of denoising random bounding box proposals, to inserting known boxes during inference.

% \subsubsection{Test time prompts for detection.}
SAM3 \cite{carion2025sam3segmentconcepts} is an iteration on the Segment Anything foundation model formula of multimodal, promptable segmentation \cite{kirillovSegmentAnything2023}. SAM3 provides state-of-the-art general purpose image segmentation, but also introduces the concept of \textit{exemplars}, test time refinement using prompts, such as bounding boxes, points or text captions describing the desired objects. These prompts are not required for inference, unlike conditioning, and not necessarily supplied during training, unlike privileged information \cite{lambertDeepLearningPrivileged2018}. SAM3 builds on the foundation of DETR, representing prompts and exemplars as embedded query representations, using a sophisticated prompt encoder, enabling joint representation of several prompt modalities. The disadvantage of this approach is that this representation is learnt. While SAM3 has a state-of-the-art data engine \cite{carion2025sam3segmentconcepts}, DETR and its queries suffer from slow learning \cite{daiDynamicDETREndtoEnd2021}.

\section{Methodology}
\begin{figure}[t]
  \centering
  \input{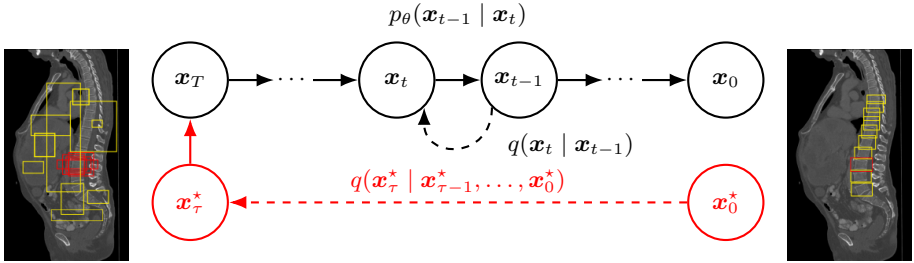}
  \caption{\textit{Model overview.} Diffusion of boxes $\bx$ through a forward process $q$ and denoising process $p_\theta$.  Known exemplars $\b x_0^\star$ are replicated $n^\star$ times, forward diffused $\tau$ steps and inserted into $\b x_T$. The joint set of proposals is denoised normally by the backward process.}
  \label{fig:diffusion}
\end{figure}
\subsection{Preliminaries}
Diffusion models, such as DiffusionDet \cite{chenDiffusionDetDiffusionModel2023}, are probabilistic models defining a Markovian process of noising and denoising of samples, see Figure~\ref{fig:diffusion}. A forward noising process $q$ gradually perturbs data $\b x_0\sim q(\b x_0)$ into noise, and a backwards model $p_\theta$ is trained to reverse the noising process. In denoising diffusion probabilistic models (DDPMs)~\cite{hoDenoisingDiffusionProbabilistic2020}, a closed form of the forward process is
\begin{equation}
    \b x_t=\sqrt{\bar\alpha_t}\b x_0+\sqrt{1-\bar\alpha_t}\,\epsilon,\quad \epsilon\sim\mathcal{N}(\b 0,\b 1),
\end{equation}
implying a forward process $q(\b x_t\mid \b x_0)=\mathcal{N}(\sqrt{\bar\alpha_t}\,\b x_0,\,(1-\bar\alpha_t)\b 1)$, where $\{\beta_t\}$ is a variance schedule \cite{hoDenoisingDiffusionProbabilistic2020} and $\alpha_t=1-\beta_t$, $\bar\alpha_t=\prod_{s=1}^t\alpha_s$.
% \begin{equation}
%   q(x_t\mid x_{t-1})=\mathcal{N}(\sqrt{1-\beta_t}\,x_{t-1},\,\beta_t),\quad t=1,\dots,T,
% \end{equation}
The reverse process is parametrized as neural networks $\mu_\theta(\b x_t,t), \mathrm\Sigma_\theta$ predicting the noise $\epsilon$:
\begin{equation}
  p_\theta(\b x_{t-1}\mid \b x_t)=\mathcal{N}(\mu_\theta(\b x_t,t),\,\mathrm\Sigma_\theta),
\end{equation}
given the sample and timestep $t$.

In DiffusionDet, and our work, a distribution of bounding box coordinates $\b x_T\in\mathbb R^4$ is diffused to learn a proposal distribution $\b x_0$ of a Sparse R-CNN detection head, conditional on an image $I$. The detection head refines the proposals using self-attention, modelling the the spatial relations of objects in the image. For each step of the sampling process $q$, refined proposals are filtered by score, with low-confidence boxes being replaced with new, random boxes, in a process known as \textit{box renewal}.
% and are typically trained by predicting noise, i.e., learning $\epsilon_\theta(x_t,t)$ such that $\epsilon_\theta(x_t,t)\approx\epsilon$.

% Denoising diffusion implicit models (DDIMs)~\cite{songDenoisingDiffusionImplicit2022} use the same trained $\epsilon_\theta$ but alter the reverse dynamics. 
% Writing the predicted clean sample as
% \begin{equation}
%   \hat{\boldsymbol{x}}_0(x_t,t)=\frac{1}{\sqrt{\bar\alpha_t}}\Bigl(\b x_t-\sqrt{1-\bar\alpha_t}\,\epsilon_\theta(\b x_t,t)\Bigr),
% \end{equation}
% DDIM instead updates samples from $t$ to $t-1$ via
% \begin{equation}
%   \b x_{t-1}=\sqrt{\bar\alpha_{t-1}}\,\hat{x}_0(\b x_t,t)+\sqrt{1-\bar\alpha_{t-1}-\sigma_t^2}\,\epsilon_\theta(\b x_t,t)+\sigma_t \epsilon,\quad \epsilon\sim\mathcal{N}(0,1),
% \end{equation}
% where $\sigma_t$ controls stochasticity (with $\sigma_t=0$ giving deterministic sampling) and enables using fewer denoising steps in practice.

\subsection{Proposed framework}
\subsubsection{Diffusion of object exemplars.} 
In this work, we aim to inform inference of unknown objects $\b x_0\in\mathbb R^4$ in an image $I$ using known exemplars $\b x_0^\star\in\mathbb R^4$ by employing the framework introduced by DiffusionDet. We parametrize the bounding boxes with class $y$ as $\b x=(c_i, c_j,w,h)$, where $(c_i,c_j)$ and $(w,h)$ are the centre point and dimensions of the box, respectively. Given $N^\star$ exemplars $\b x_0^\star$ at inference, we may use them as priors for the forward sampling distribution 
\begin{equation}
    q(\b x^\star_t|\b x^\star_0) =\mathcal{N}(\sqrt{\bar\alpha_t}\,\b x_0,\,(1-\bar\alpha_t)\b 1),\quad t=1,\dots,T.
\end{equation}
In practice, given $T$ diffusion steps, $n^\star$ copies of in total $N^\star$ different $\b x^\star_0$ are forward diffused and inserted into the set of $n$ random, normal distributed box proposals $\b x_T$, see Fig.~\ref{fig:diffusion}. We may retain more information of the exemplars by diffusing them for potentially fewer steps, $\tau\leq T$. The set of proposals is then
\begin{equation}
\b x_T =
\begin{cases}
    \,\epsilon, \quad \epsilon\sim\mathcal{N}(\b 0,\b 1), &i=1,...,n \\[0.5em]
    \,\sqrt{\bar\alpha_\tau}\b x_0^\star+\sqrt{1-\bar\alpha_\tau}\,\epsilon,\quad \epsilon\sim\mathcal{N}(\b 0,\b 1),&i=n+1, ...n+N^\star n^\star.
\end{cases}
\end{equation}
The set of proposals, containing random boxes and known exemplars, are denoised in the diffusion backward process, and refined by a detection head. Both random proposals and exemplars are subject to box renewal.

\subsubsection{Exemplar-anchored uncertainty.}
We also show how to use exemplars in DiffusionDet to estimate predictive uncertainty. 
We create an ensemble of diffusion processes, yielding a distribution of predictions. The predictions are then matched with the exemplar at an intersection-over-union (IoU) threshold. Given predicted bounding box corners $\b x$ matched with an exemplar $\b x^\star$, the confidence interval (CI) may be given by the Mahalanobis distance $d$:
\begin{equation}
    d^2 = (\b x - \b \mu)\mathrm\Sigma^{-1}(\b x - \b \mu),
\end{equation}
where $\b \mu$ and $\mathrm\Sigma$ are the means and covariance matrix of the matched boxes, respectively, and $d^2$ is $\chi^2$-distributed. Assuming that the CI is the same for all predictions, this should produce a lower bound estimate on the remaining CIs.

% \subsubsection{Uncertainty quantification.} 
\begin{figure}[t]
  \centering
  \begin{tikzpicture}
\begin{axis}[
    width=12cm,
    height=4.5cm,
    xlabel={\textit{number of exemplars,} $N^\star$},
    ylabel={AP},
    grid=both,
    xmin=0,
    xmax=3,
    ymin=33.5,
    ymax=36.5,
    legend pos=north west,
    enlargelimits=true,
]

% -------------------------------------------------------------------------
% Shaded confidence band (mean ± std)
% -------------------------------------------------------------------------
% \addplot[name path=upper, draw=none, forget plot]
%     table[x=n_known, y expr=\thisrow{AP_mean} + \thisrow{AP_std}] {              
% n_known AP_mean AP_std
% 0	34.44	2.15
% 1	35.28	2
% 2	35.78	1.86
% 3	35.24	2.11
%     };

% \addplot[name path=lower, draw=none, forget plot]
%     table[x=n_known, y expr=\thisrow{AP_mean} - \thisrow{AP_std}] {
% n_known AP_mean AP_std
% 0	34.44	2.15
% 1	35.28	2
% 2	35.78	1.86
% 3	35.24	2.11
%     };

% \addplot[blue!30, fill opacity=0.4, mark=none, forget plot] 
%     fill between[of=upper and lower];

% -------------------------------------------------------------------------
% Mean line
% -------------------------------------------------------------------------
\addplot[
    blue,
    thick,
    mark=*,
]
table[x=n_known, y=AP_mean] {
n_known AP_mean AP_std
0	34.44	2.15
1	35.28	2
2	35.78	1.86
3	35.24	2.11
};

\addplot[dashed, thick] coordinates {(-1, 34.44) (5, 34.44)};

\legend{AP, \textit{baseline} AP}

% \addlegendimage{dashed, gray, mark=none}
% \addlegendentry{\textit{baseline}}

\end{axis}
\begin{axis}[
    width=12cm,
    height=4.5cm,
    xmin=0,
    xmax=3,
    axis x line=none,
    axis y line*=right,
    ymin=61.5,
    ymax=65.5,
    ylabel={AR},
    yticklabel style={/pgf/number format/fixed},
    legend pos=south east,
    enlargelimits=true,
]

% -------------------------------------------------------------------------
% Mean line
% -------------------------------------------------------------------------
\addplot[
    red,
    thick,
    mark=square,
]
table[x=n_known, y=AP_mean] {
n_known AP_mean AP_std
0	62	1.7
1	64.14	0.94
2	64.84	2.2
3	64.86	1.49
};

\addplot[dotted, thick] coordinates {(-1, 62.00) (5, 62.00)};

\legend{AR, \textit{baseline} AR}

\end{axis}
\end{tikzpicture}
  \caption{\textit{Impact of number of exemplars}. Average recall increases with number of exemplars $N^\star$, while average precision decreases beyond $N^\star=2$, on the DENTEX dataset, where the number of labels is approx. 3.5.}
  \label{fig:n_known_ablation}
\end{figure}
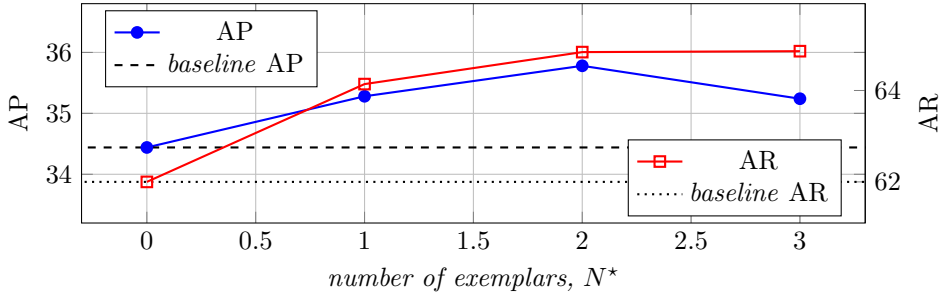
\section{Experiments and results}
\subsubsection{Data.}
We evaluate or model on the DENTEX dataset \cite{hamamciDiffusionBasedHierarchicalMultiLabel2023} of panoramic dental X-rays, testing our approach on the task of diseased teeth detection and diagnosis.

Secondly, we test model sensitivity to the randomness of bounding boxes in the images, using three additional datasets, each with increasing spatial randomness compared to DENTEX. The VerSe dataset \cite{sekuboyinaVerSeVertebraeLabelling2021} of spinal CTs for vertebra detection, originally intended for 3D segmentation, where segmentation masks are converted into bounding boxes at the middle sagittal slice. Lateral radiography of the vertebra are common in screening for e.g. osteoporotic fractures \cite{waahlstrand2024explainable,Dong2023}, and VerSe constitutes an open, accessible option for this use-case. 

Lastly, we compare with two datasets with nearly random spatial distribution of objects: the liver disease dataset \cite{liver-disease_dataset} containing annotations of anomalous liver tissues, and the TXL-PBC dataset \cite{gan2025curated}, a curated datasets with images of multiple classes of blood cells. 

\subsubsection{Model implementation and training.}
We implement standard DiffusionDet model configurations, trained on the DENTEX, VerSe, Liver and TXL-PBC datasets, with $n=500$ box proposals and early stopping. The number of exemplar copies $n^\star$ provides a compromise between exploration of random proposals, and exploitation of the provided exemplars. We find that the proportion is highly dataset dependent, but that a $n^\star=200$ generally performs well for $n=300$ random box proposals. We sample in total 10 steps from $t=1000$ to $0$. Lastly, the number of exemplar noising steps is found to be low, $\tau\approx10$.

For all datasets, we pool the training and validation sets (if available) and perform 5-fold cross-validation, evaluating on holdout test sets. The splits will be made openly available. Models are evaluated in terms of average precision (AP) at various thresholds of IoU, $\ap=\apat{[0.5,0.95]}$, $\apat{50}$, and average recall (AR). Exemplars are randomly selected and then completely removed from evaluation, ensuring that they do not inflate results.

\subsection{Ablations}

\begin{table}[t]
  \centering
  \caption{\textit{Annotation randomness} Comparison across datasets with varying spatial entropy $H$. Metrics reported as the mean on test set over cross-validation folds. Changes compared to the baseline $N^\star=0$.}
  \label{tab:dataset_model_ap_ar_transposed}
  \fontsize{9}{10.8}\selectfont
  \setlength{\tabcolsep}{3.0pt}
  \renewcommand{\arraystretch}{1}
  \begin{tabular}{c l | *{4}{>{\fontsize{9}{9.5}\selectfont}c}}
    \toprule
    & Dataset & DENTEX\,\cite{hamamciDiffusionBasedHierarchicalMultiLabel2023} & VerSe\,\cite{sekuboyinaVerSeVertebraeLabelling2021} & Liver\,\cite{liver-disease_dataset} & TXL-PBC\,\cite{gan2025curated} \\
    \midrule
     & \multirow{2}{*}{entropy} & $H=0.66$ & $H=0.80$ & $H=0.98$ & $H=0.98$  \\
     & &
    \multicolumn{4}{@{}c@{}}{%
      \begin{tikzpicture}
        \draw[->, line width=0.4pt] (0,0) -- node[midway, fill=white] {$H$}(6cm,0) ;
      \end{tikzpicture}%
    }\\
    \midrule
    \multirow{3}{*}{$N^\star=0$} & AP & 34.44 & 48.68 & 26.74 & 83.94 \\
    & $\mathrm{AP}_{50}$ & 51.80 & 62.30 & 29.72 & 98.54 \\
    & AR & 62.00 & 76.74 & 48.24 & 89.14 \\
    \midrule
    \multirow{3}{*}{$N^\star=1$} & AP & 35.36\added{1.0} & 49.06\added{0.4} & 26.98\added{0.2} & 84.12\added{0.2} \\
    & $\mathrm{AP}_{50}$ & 53.56\added{1.4} & 62.76\added{0.4} & 29.90\added{0.1} & 97.52\lost{0.0} \\
    & AR & 64.14\added{2.1} & 76.60\lost{0.1} & 47.96\lost{0.3} & 89.22\added{0.0}\\
    % \multirow{3}{*}{$N^\star=1$} & AP & 35.36\added{0.9} & 49.06\added{0.4} & 26.98\added{0.2} & 84.12\added{0.2} \\
    % & $\mathrm{AP}_{50}$ & 53.56\added{1.8} & 62.76\added{0.4} & 29.90\added{0.1} & 97.52\lost{0.0} \\
    % & AR & 64.14\added{2.1} & 76.60\lost{0.1} & 47.96\lost{0.3} & 89.22\added{0.0}\\
    % \midrule
    % \multirow{3}{*}{$N^\star=2$} & AP & 35.78\added{1.3} & 48.74\added{0.0} & 26.78\added{0.0} & 84.08\added{0.1} \\
    % & $\mathrm{AP}_{50}$ & 54.34\added{2.5} & 63.00\added{0.7} & 29.64\lost{0.1} & 97.52\added{0.0} \\
    % & AR & 64.84\added{2.7} & 76.22\lost{0.5} & 47.84\lost{0.4} & 89.22\added{0.1} \\
    \bottomrule
  \end{tabular}
\end{table}

\subsubsection{Number of different exemplars.}
If a single exemplar is informative about the locations of other objects in an image, it is reasonable to expect that providing multiple exemplars should be further beneficial. We ablate this effect on the DENTEX dataset, bearing in mind that the average number of ground truth boxes in the dataset is $3.5$, providing an upper limit to the number of exemplars. Figure~\ref{fig:n_known_ablation} shows that while both average recall and precision increase with the number of exemplars, AP falls off after $N^\star=2$. However, since the objective is to provide minimal annotation, using more exemplars is unlikely in practice.
% \begin{table}[t]
%   \centering
%   \caption{\textit{Robustness to noise.} The model shows robustness to adding noise with scale $\sigma$ to the input exemplar ($N^\star=1)$, changes relative to baseline $N^\star=0$ on the DENTEX dataset.}
%   \label{tab:nstar}
%   \fontsize{9}{10.8}\selectfont
%   \setlength{\tabcolsep}{6pt}
%   \renewcommand{\arraystretch}{1.15}
%   \begin{tabular}{l | c c c}
%     \toprule
%      &
%     \multicolumn{3}{c}{number of different exemplars} \\
%      &  $N^\star=1$ & $N^\star=2$ & $N^\star=3$  \\
%     \midrule
%      AP & --- & --- & --- \\
%      $\mathrm{AP}_{50}$ & --- & --- & --- \\
%      AR & --- & --- & --- \\
%     \bottomrule
%   \end{tabular}
% \end{table}

\subsubsection{Annotation randomness.}
Since the Sparse R-CNN detector head uses self-attention between bounding boxes, our model likely benefits from images with a clear spatial structure (i.e., bounding boxes not appearing completely randomly in relation to each other). We estimate this effect in different datasets by computing the entropy $H$ for bounding box coordinates, where a low value indicates less randomness. Table~\ref{tab:dataset_model_ap_ar_transposed} reports results on the DENTEX, VerSe, Liver and TXL-PBC, the former two containing images of teeth and vertebrae arranged consistently across images, with $H=0.66$ and $H=0.80$. Liver and TXL-PBC both use cellular data, with more random box dispersion, such that $H=0.98$. As expected, exemplars provide greater utility for the more structured datasets in terms of $\ap$, $\apat{50}$ and $\ar$.

\subsection{Model performance}
\subsubsection{Exemplar model comparison.}
\begin{table}[t]
  \centering
  \caption{\textit{Exemplar model comparison} for different numbers of exemplars $N^\star\in\{0,1,2\}$ in terms of average recall on the DENTEX dataset, the text prompt supplied to SAM3 being ``\textit{do not segment healthy teeth}''. Statistical significance ($^\ast$) from baseline ($N^\star=0)$ using two-sided $t$-test. }
  \label{tab:sam3_comparison}
  \fontsize{9}{10.8}\selectfont
  \setlength{\tabcolsep}{5pt}
  \renewcommand{\arraystretch}{1}

\begin{tabular}{l l | c c c c}
    \toprule
    & & SAM3 & SAM3 \textit{w. text prompt} & Sparse R-CNN & \textit{ours} \\
    \midrule
    \multirow{3}{*}{$N^\star=0$} & AP & --- & --- & $28.24\pm 2.60$ & $34.44\pm 2.15$ \\
    & $\mathrm{AP}_{50}$ & --- & --- & $43.30\pm 2.96$ & $51.80\pm 2.36$ \\
    & AR & --- & --- & $68.88\pm0.86$ & $62.00\pm 1.70$ \\
    \midrule
    \multirow{3}{*}{$N^\star=1$} & AP & 4.74 & 5.05 & $28.24\pm 2.60$ &  \boldmath\textbf{$35.36\pm 2.08$} \\
    & $\mathrm{AP}_{50}$ & 13.02 & 14.28 & $43.30\pm 2.96$ &  \boldmath\textbf{$53.56\pm3.07$} \\
    & AR & 39.41 & 37.57 &  \boldmath\textbf{$68.88\pm0.86$} & $64.14\pm0.94$ \\
    \midrule
    \multirow{3}{*}{$N^\star=2$} & AP & 3.97 & 4.40 & $28.26\pm 2.58$ &  \boldmath\textbf{$35.78\pm 1.86$} \\
    & $\mathrm{AP}_{50}$ & 9.20 & 10.51 &  $43.32\pm 2.94$ &  \boldmath\textbf{$54.34^\ast\pm 2.00$} \\
    & AR & 45.53 & 43.30 &  \boldmath\textbf{$68.86\pm0.79$} & $64.84^\ast\pm2.20$ \\
    \bottomrule
  \end{tabular}
\end{table}
\begin{table}[t]
  \centering
  \caption{\textit{Robustness to noise.} The model shows robustness to adding noise with scale $\sigma$ to the input exemplar ($N^\star=1)$, changes relative to $N^\star=0$ on DENTEX.}
  \label{tab:noise}
  \fontsize{9}{10.8}\selectfont
  \setlength{\tabcolsep}{6pt}
  \renewcommand{\arraystretch}{1}
  \begin{tabular}{l | c c c}
    \toprule
    $N^\star=1$ &  $\sigma=1$ & $\sigma=5$ & $\sigma=10$  \\
    \midrule
     AP & 35.00\added{0.6} & 34.80\added{0.4} & 34.50\added{0.0} \\
     $\mathrm{AP}_{50}$ &  52.92\added{0.9} & 52.62\added{0.6} & 52.12\added{0.3} \\
     AR & 63.56\added{1.5} & 63.16\added{1.1} & 63.66\added{1.6}\\
    \bottomrule
  \end{tabular}
\end{table}

% For exhaustion, we compare our results with SAM3 and a naïve approach using Sparse R-CNN. However, the intention of exemplars in SAM3 differs slightly from our objective, where the former uses exemplars to zero-shot segment all similar objects, without awareness of object classes. If detected similar objects do not have the same class, a disambiguating negative exemplar or text prompt is required, alternatively training a new classifier. As such, SAM3 will produce numerous false positives using its exemplars, lowering the average precision. For fairness, Table~\ref{tab:sam3_comparison} also compares SAM3 with an additional text prompt, ``\textit{do not segment healthy teeth}'' which was found to be the best performing one. SAM3 is capable of segmentation without a prompt ($N^\star=0)$, but this results in an exhaustive segmentation without control of the targets, leading to an unfair comparison. Table~\ref{tab:sam3_comparison}

For exhaustion, we compare our results with SAM3. Unlike our objective, SAM3 uses exemplars to zero-shot segment all similar objects without class awareness. When similar-looking objects belong to different classes, SAM3 requires a negative exemplar or a text prompt. Consequently, SAM3 often produces many false positives when relying solely on exemplars, reducing average precision. For fairness, we also include SAM3 with a text prompt, “\textit{do not segment healthy teeth},” which yielded the best performance. We do not compare SAM3 without prompts ($N^\star=0$), since the resulting segmentation map would not be related to the DENTEX targets.

We also compare with a naïve Sparse R-CNN approach, based on the realization that DiffusionDet proposals are processed by a Sparse R-CNN detection head as embedded object queries. As such, we compare with simply inserting the exemplars as queries to the detection head. Figure~\ref{fig:preds} shows a sample image with predictions given an exemplar for our model. Table~\ref{tab:sam3_comparison} shows the quantitative results for $N^\star=1,2$ exemplars on DENTEX data. Sparse R-CNN, although it performs well in terms of recall, does not improve at all when manipulating the queries. The SAM3 methods show large improvements in AR for additional exemplars, but cannot compete at the level of a dedicated detectors. 

\subsubsection{Robustness to noise.}
A main contribution of exemplar diffusion is the possibility of adding minimal, informative annotation to improve predictions. However, since any annotation is costly, high-quality, expert-annotated exemplars may be a limiting factor. In Table~\ref{tab:noise} we simulate non-expert annotators as Gaussian noise in the exemplar boxes, and demonstrate that the model still benefits from exemplars across a range of noise levels ($\sigma\in[1,10]$).

\subsubsection{Uncertainty calibration.}
\begin{figure}[t!]
    \centering
    \input{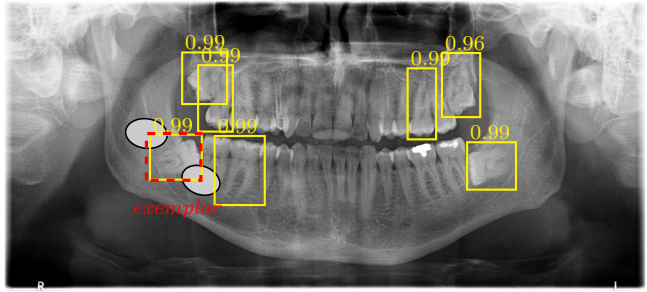}
    \caption{\textit{Results visualization}. Sample image of teeth detected as diseased by our model (\textit{yellow}) using a known exemplar (\textit{red, dashed}). Ellipses indicate an estimated $95\%$ confidence interval for the box coordinates.}
    \label{fig:preds}
\end{figure}

Figure~\ref{fig:preds} shows an example of $95\%$ CIs as shaded ellipses, based on matched predictions with the exemplar (\textit{red}) at a level of IoU$=0.5$, using an ensemble of 100 diffusion processes. We assume that this is likely a lower bound for the uncertainty of other detections, but test the hypothesis in terms of the calibration on the test set. Calculating the number of predicted boxes that fall in the corresponding CIs yields a median of $83\%$, or a calibration error of 12. This indicates that the uncertainty is likely underestimated, but potentially a fast alternative to explicit modelling of the uncertainties.

\section{Discussion and conclusion}
We present \textit{exemplar diffusion}, a training-free way to improve medical object detection by injecting a small number of known bounding boxes (exemplars) at test time. Across datasets, exemplars improved both precision and recall when anatomy exhibits stable spatial structure, while yielding smaller gains when objects are more randomly dispersed. This suggests that a confirmed location can act as a strong spatial prior that helps disambiguate nearby proposals, but that the same cue is less informative in settings with higher spatial entropy.

Exemplar diffusion can utilize \emph{opportunistic} annotations, existing in medical datasets with missing labels, into better inference without retraining or learning prompt representations. The method operates directly in image coordinates, making it easy to integrate into existing DiffusionDet models. We demonstrate it to be robust to noisy exemplars, indicating that coarse, non-expert input may still be useful in practice. Finally, we showed that exemplars can be used to obtain a lightweight, approximate uncertainty estimate for diffusion-based detectors; while calibration remains imperfect, it may still serve as a fast diagnostic.

While this paper has explored the specific use case of DiffusionDet bounding box detection, the model is easily extendable into e.g. keypoint regression. Encouraging future directions however, should include generalizing the framework into segmentation and 3D medical imaging and implementing more visual information when exemplar locations are weakly informative.

% \begin{credits}
%   \subsubsection{\ackname} The authors would like to acknowledge 
%   \subsubsection{\discintname}
%   The authors declare no competing interests.
% \end{credits}

\newpage
\bibliographystyle{splncs04}
\bibliography{bibliography}

\end{document}